\documentclass[conference]{IEEEtran}
\IEEEoverridecommandlockouts
\usepackage{cite}
\usepackage{amsmath,amssymb,amsfonts}
\usepackage{algorithmic}
\usepackage{graphicx}
\usepackage{textcomp}
\usepackage{xcolor}
\def\BibTeX{{\rm B\kern-.05em{\sc i\kern-.025em b}\kern-.08em
    T\kern-.1667em\lower.7ex\hbox{E}\kern-.125emX}}

\usepackage{booktabs}
\usepackage{multirow}
\usepackage{makecell}
\usepackage{array}  

\begin{document}

\title{Revisit the Algorithm Selection Problem for TSP with Spatial Information Enhanced Graph Neural Networks
}

\author{\IEEEauthorblockN{1\textsuperscript{st} Ya Song}
\IEEEauthorblockA{
\textit{Eindhoven University of Technology}\\
5600 MB Eindhoven, Netherlands \\
y.song@tue.nl}
\and
\IEEEauthorblockN{2\textsuperscript{st} Laurens Bliek}
\IEEEauthorblockA{
\textit{Eindhoven University of Technology}\\
5600 MB Eindhoven, Netherlands \\
l.bliek@tue.nl}
\and
\IEEEauthorblockN{3\textsuperscript{st} Yingqian Zhang}
\IEEEauthorblockA{
\textit{Eindhoven University of Technology}\\
5600 MB Eindhoven, Netherlands \\
yqzhang@tue.nl}
}

\maketitle

\begin{abstract}
Algorithm selection is a well-known problem where researchers investigate how to construct useful features representing the problem instances and then apply feature-based machine learning models to predict which algorithm works best with the given instance. However, even for simple optimization problems such as Euclidean Traveling Salesman Problem (TSP), there lacks a general and effective feature representation for problem instances. 

The important features of TSP are relatively well understood in the literature, based on extensive domain knowledge and post-analysis of the solutions. In recent years, Convolutional
Neural Network (CNN) has become a popular approach to select algorithms for TSP. Compared to traditional feature-based machine learning models, CNN has an automatic feature-learning ability and demands less domain expertise. However, it is still required to generate intermediate representations, i.e., multiple images to represent TSP instances
first. 

In this paper, we revisit the algorithm selection problem for TSP, and propose a novel Graph Neural Network
(GNN), called GINES. GINES takes the coordinates of cities and distances between cities as input. It is composed of a new message-passing mechanism and a local neighborhood feature extractor to learn spatial information of TSP instances. We evaluate GINES on two benchmark datasets. The results show that GINES outperforms CNN and the original GINE models. It is better than the traditional handcrafted feature-based approach on one dataset. The code and dataset will be released in the final version of this paper.
\end{abstract}


\section{Introduction}
The Euclidean Traveling Salesman Problem (TSP) is one of the most intensely studied NP-hard combinatorial optimization problems. It relates to many real-world applications and has significant theoretical value. TSP can be described as follows. Given a list of cities with known positions, find the shortest route to visit each city and return to the origin city. Researchers have developed various exact, heuristic, and learning-based algorithms to solve this routing problem~\cite{joshi2019efficient}. As these algorithms' performance is highly variable depending on the characteristics of the problem instances, selecting algorithms for each instance helps to improve the overall efficiency~\cite{huerta2022improving}. 
The algorithm selection problem was proposed
in~\cite{rice1976algorithm}, 
and developed further in~\cite{smith2011discovering,smith2014towards}, where the authors consider the algorithm selection as a classification problem that identifies the mapping from Problem Space to Algorithm Space~\cite{kotthoff2016algorithm}. Traditionally, domain experts design a group of features~\cite{hutter2014algorithm,bosseksalesperson,pihera2014application} that can represent the characteristics of TSP instances well. Then, one can train a machine learning classifier to be the selector using these features. This feature-based method has several potential limitations:
high requirement for domain knowledge, insufficient expressiveness of the features~\cite{matejka2017same}, and required feature selection process~\cite{seiler2020deep}. Handcrafted features are not effective when being directly transferred to represent instances of other optimization problems. For complex optimization problems that are much less studied than TSP, it is hard for humans to design good features to represent instances.  

Deep learning models, especially Convolutional Neural Networks (CNN), have recently been applied to select TSP algorithms. By employing images to represent TSP instances, the algorithm selection problem is transformed into a computer vision challenge. Since CNN has sufficient automatic feature learning capability, this approach no longer requires handcrafted 
features. 
In~\cite{seiler2020deep}, the authors generate three images: a point image, a Minimum Spanning Tree (MST) image, and a k-Nearest-Neighbor-Graph (kNNG) image to represent each TSP instance. Then they apply an 8-Layer CNN architecture to predict which algorithm is better. In~\cite{zhao2021towards}, researchers use a gridding method to transform TSP instances into density maps, and then apply Residual Networks (ResNet)~\cite{he2016deep} to do the classification. In~\cite{huerta2022improving}, a similar gridding approach is used to generate images, and then a 3-Layer CNN model is designed to predict algorithms' temporal performance at different time steps.

Although the experimental results in~\cite{huerta2022improving,seiler2020deep,zhao2021towards} show CNN can outperform traditional feature-based machine learning models in the algorithm selection task for TSP, this approach still has the following main drawbacks: (1)
 \textit{
  Need to generate intermediate representations}.  Similar to feature-based methods, the instances' intermediate representations, in this case, the images, need to be generated as the inputs of CNN. It is usually a tedious process to transform TSP instances into images. In~\cite{seiler2020deep}, generating MST and kNNG images for each instance requires time-consuming calculations. When applying the gridding method to obtain images, the authors perform several up-scaling operations to improve the resolution~\cite{zhao2021towards}. Besides, data augmentation techniques, like random rotation/flipping, are widely used to enhance CNN's generalization ability~\cite{huerta2022improving,seiler2020deep,zhao2021towards}. As a result, multiple images must be generated to represent one TSP instance.
(2)  \textit{Introduce problem-irrelevant parameters}. In~\cite{seiler2020deep}, authors use solid dots to represent cities and solid lines to connect cities in MST and kNNG images. The dot size and line width are irrelevant to the properties of the TSP instance. Similarly, when applying gridding methods to generate images, the key parameter we need to set is the image size or the number of grids~\cite{zhao2021towards}. Adding these parameters increases the input data's complexity and the effort required for parameter tuning.
 (3) \textit{Potentially lose problem-relevant information}. In the image generation procedure, the TSP instance is divided into multiple grids, with the value for each grid representing the number of cities that fall into it~\cite{huerta2022improving,zhao2021towards}. After gridding, portions of the instance's local structure will be lost. In addition, \cite{huerta2022improving} sets a maximum number for the value of grids, leading to more information distortion. 
(4) \textit{Hard to generalize to other routing problems}. 
  The gridding methods can be applied to convert TSP instances to images since cities are in 2D Euclidean space. However, for many variants of TSP problems, such as 
the Asymmetric Traveling Salesman Problem (ATSP) and the Capacitated Vehicle Routing Problem (VRP), generating images to represent problem instances is not straightforward and could be very challenging. In such cases, a graph with assigned node/edge features could be a better representation form. 

To remedy the above issues,  we propose an enhanced Graph Neural Network (GNN) named GINES to solve algorithm selection problems for TSP. Our main contributions are:

\begin{itemize}
\item We are the first to successfully design a GNN to learn the representation of TSP instances for algorithm selection, outperforming the existing feature-based or CNN-based approaches. 
\item The proposed model merely takes the coordinates of cities and the distance between them as inputs. We show there is no need to design and generate intermediate representations, such as handcrafted features or images, for TSP instances.
\item The adopted graph representation methodology has few parameter settings, and the experimental results show it can retain accurate information about the original TSP instances.
\item The proposed model is able to capture local features with multiple scales by aggregating information from the neighborhood nodes. Its robust performance is demonstrated on two public TSP datasets, compared with several existing approaches.
\item The proposed model can easily generalize to other complex routing problems by adding node features or modifying distance metrics.
\end{itemize}

The rest of the paper is organized as follows.  Section 2 introduces the background and related works. Section 3 presents the proposed GINES. Section 4 shows the experimental results of GINES. We conclude in Section 5. 
\section{Background and Related Work}

\subsection{Algorithm selection for optimization problems}
The No Free Lunch (NFL) theorem states that no algorithm can outperform others on all optimization problems. Researchers have been investigating algorithm selection problems to improve overall solving performance~\cite{kerschke2019automated}. Most researchers focus on designing features for problem instances and solving algorithm selection by traditional feature-based machine learning models. The collection of features for 
classical  optimization problems like Satisfiability Problem~\cite{hoos2014claspfolio}, AI planning~\cite{fawcett2014improved}, Knapsack Problem~\cite{huerta2020anytime},  TSP~\cite{hutter2014algorithm,bosseksalesperson,pihera2014application}, and VRP~\cite{mayer2018simulation,rasku2019toward} have been well designed. These features are restricted to specific problems and usually need great efforts to be generated.

Deep learning has been shown to perform various classification/regression tasks effectively. 
In addition to the algorithm selection models using CNN for TSP mentioned above~\cite{huerta2022improving,seiler2020deep,zhao2021towards}, researchers have proposed a few feature-free algorithm selection models for other optimization problems. By generating images from the text documents for SAT problem instances, CNN can be applied to selecting algorithms~\cite{loreggia2016deep}. In~\cite{he2020black}, researchers sample landscape information from instances and transform it into images, then apply CNN to select algorithms for Black-Box Optimization Benchmarking (BBOB) function instances. The authors of ~\cite{alissa2023automated} treat online 1D Bin-Packing Problem instances as sequence data and apply Long Short-Term Memory (LSTM) to predict heuristic algorithms' performance. In the feature-free algorithm selection field, instances are usually converted to images or sequences, and graph representations are seldom used.

\subsection{Hardness prediction for optimization problems}
Instance hardness prediction is a research topic closely related to algorithm selection. The purpose of the hardness prediction is to assess whether the problem instance is easy or difficult to solve using a specific algorithm. Researchers have studied where are the hard optimization problem instances, 
especially the hard TSP~\cite{hougardy2021hard} or Knapsack instances~\cite{pisinger2005hard,jooken2022new}. Similar to algorithm selection, the main research idea is to identify key attributes which correlate with hardness levels. The authors of ~\cite{cheeseman1991really} point out that the Standard Deviation (SD) of the distance matrix is highly relevant to TSP instance hardness. In ~\cite{smith2010understanding}, the same features used in the TSP algorithm selection are applied to predict TSP instance hardness for local search algorithms. Some other complex features, such as the highest edge features~\cite{mersmann2012local}, the clustering features~\cite{kromer2018evaluation}, Weibull distribution of distances~\cite{cardenas2018creating},  are proposed to assess the TSP hardness to some heuristic algorithms such as Ant Colony Optimization (ACO). In ~\cite{cricsan2021randomness}, researchers find that the regularity of the TSP structure can indicate the TSP hardness to ACO, but this type of feature cannot predict the hardness to the local search Lin-Kernighan algorithm. It implies that the valuable features may vary across algorithms. Researchers commonly apply traditional feature-based machine learning models in this research area, and no deep learning models have been employed to our knowledge.

\subsection{GNN for TSP}
The TSP instance can be naturally expressed by a graph $ G = (V, E) $, where $ V = \{v_{1},v_{2},...,v_{n}\} $ is a group of cities, and $ E = \{\langle v_{i},v_{j}\rangle: v_{i},v_{j}\in V\} $ is a set of paths between cities. 
Therefore, there exist several research lines for applying GNNs to TSP.
\paragraph{GNN for TSP solving} GNN has been successfully applied in learning-based TSP algorithms, either in the manner of reinforcement learning or supervised learning~\cite{vesselinova2020learning}. In reinforcement learning methods, researchers use graph embedding networks such as \textit{structure2vec}~\cite{khalil2017learning} and Graph Pointer Networks (GPN)~\cite{ma2019combinatorial} to represent the current policy and apply Deep Q-Learning (DQN) to update it. To tackle larger graphs, \cite{manchanda2019learning} introduces a two-stage learning procedure that firstly trains a Graph Convolutional Network (GCN)~\cite{kipf2016semi} to predict node qualities and prune some of them before taking the next action. In supervised learning methods, GNN models are commonly used as the Encoder tool~\cite{nowak2016divide,sultana2022learning} in the upgraded version of Pointer Network~\cite{vinyals2015pointer}, a sequence-to-sequence architecture. 
\paragraph{GNN for TSP search space reduction} Search space reduction for TSP instances is another GNN-related research task, and it can be viewed as an edge classification problem. Suppose a learned model can predict which edges in the TSP instance graph are likely to be included in the optimal solution. In that case, we can reduce the search space and improve computational efficiency in the following searching procedure~\cite{fitzpatrick2021learning}.  In~\cite{dwivedi2020benchmarking}, authors have designed a benchmark TSP dataset for edge classification. Here a TSP instance is represented as a kNNG, where node features are node coordinates and edge features are Euclidean distances between two nodes. 
Many researchers use this benchmark dataset to assess the proposed GNN architectures~\cite{zhang2022ssfg}.

\paragraph{GNN for TSP algorithm selection} 
The authors of ~\cite{zhao2021towards} investigate utilizing both CNN and GCN to select TSP algorithms and conclude that CNN performs better than GCN. The authors analyze the drawbacks of GCN, including the lack of relevant node features, the over-smoothing problem~\cite{chen2020measuring}, and high time complexity. 
To the best of our knowledge, no GNN models have been successfully applied in algorithm selection for routing problems. We aim to design a suitable GNN architecture for solving the TSP algorithm selection problem.

\section{TSP algorithm selection with GINES}
\subsection{Problem Statement}
The TSP algorithm selection problem can be defined as follows: given a TSP instance set $ I = \{I_{1}, I_{2},..., I_{l}\} $, a TSP algorithm set $ A = \{A_{1}, A_{2},..., A_{m}\} $, and a certain algorithm performance metric, the goal is to identify a per-instance mapping from $I$ to $A$ that maximizes its  performance on $I$ based on the given metric. As discussed in previous sections, the TSP instances can be represented by handcrafted features or images, which are inputs to supervised learning models such as SVM and CNN to learn this mapping.

In this work, we treat a TSP instance $I_{i}$ as a graph  $G_{i} = (V, E)$, where the node features $X_{v}$ for $v \in V$  is a vector of its $(x_{v}, y_{v})$ coordinate, the edge feature $e_{u,v}$ for $(u,v)\in E$ is the  Euclidean distance between two nodes. Here we use kNNG to represent TSP instances. We set the number of nearest nodes $k$ to 10, which is relatively small compared to other papers~\cite{joshi2019efficient, dwivedi2020benchmarking} in order to reduce the computational burden. Let $N$ be the number of cities. The node feature is a $[N, 2]$ matrix, and the matrix size of the edge feature is $[N\times10, 1]$. Given a set of TSP graphs $\{G_{1}, G_{2},..., G_{l}\} $ and their algorithm performance labels $\{y_{1}, y_{2},..., y_{l}\}$,  the task of selecting TSP algorithms can be converted to a graph-level classification task. We develop a GNN model for routing problems, called GINES, which directly takes TSP graphs as inputs for classification. Next, we will describe the architecture of this model in detail.

\subsection{GINES} Graph Isomorphism Network (GIN) is one of the most expressive GNN architectures for the graph-level classification task. Researchers have shown that
the representational power of GIN is equal to the power of the Weisfeiler Lehman graph isomorphism test, and GIN can obtain state-of-the-art performance on several graph classification benchmark datasets~\cite{xu2018powerful}. GIN uses the following formula for its neighborhood aggregation and message-passing:
\begin{equation}
\mathbf{x}^{\prime}_i =  \textrm{MLP} \left( (1 + \epsilon) \cdot\mathbf{x}_i +  \sum_{j \in \mathcal{N}(i)} \mathbf{x}_j\right) \label{XX}
\end{equation}
where $\mathbf{x}_i$ is the target node's features, 
$\mathcal{N}(i)$ denotes the neighborhood for node $i$, and $\mathbf{x}_j$ is the neighborhood nodes' features.  
$\epsilon$ indicates the significance of the target node relative to its neighborhood, with a default value of zero. $\mathbf{x}^{\prime}_i$ is the representation of node $i$ we get after applying one GIN layer. Here, often the SUM aggregator is used to aggregate information from the neighborhood, as it can better distinguish different graph structures than MEAN and MAX aggregators~\cite{xu2018powerful}. 
A drawback of the original GIN is that the edge features are not taken into account. 
Thus, the authors of~\cite{hu2019strategies} proposed GINE that can incorporate edge features in the aggregation procedure:  
\begin{equation}
\mathbf{x}^{\prime}_i =  \textrm{MLP} \left( (1 + \epsilon) \cdot\mathbf{x}_i +  \sum_{j \in \mathcal{N}(i)} \textrm{ReLU} \left( \mathbf{x}_j +  \mathbf{e}_{j,i}\right)\right) \label{XX}
\end{equation}
where $\mathbf{e}_{j,i}$ are edge features. In GINE, the neighborhood nodes' features and edge features are added together and make a ReLU transform before the SUM aggregation. With a TSP graph, the dimensions of these two features do not match. Therefore, we perform a linear transform to edge features. 

To better tackle the TSP algorithm selection problem, we make several modifications on GINE and propose a GINES (GINE with Spatial information) architecture as follows.

\paragraph{\textbf{Adopting a suitable aggregator}} Aggregators in GNNs play a crucial role in incorporating neighborhood information. Researchers have illustrated that the selection of aggregators significantly impacts GNN's representational capacity~\cite{xu2018powerful}. The widely applied aggregators are the MEAN aggregator, MAX aggregator, and SUM aggregator, and which aggregator is the best is an application-specific question. For example, the MEAN aggregator used in GCN can help to capture the nodes distribution in graphs, and it may perform well if the distributional information in the graph is more relevant to the studied task~\cite{xu2018powerful}. The MAX aggregator is beneficial for identifying representative nodes, and thus for some vision tasks like point clouds classification, the MAX aggregator is a better choice~\cite{qi2017pointnet++}. The SUM aggregator enables the learning of structural graph properties, which is the default setting of GIN. 

With post-analysis, 
researchers have shown that the standard deviation (SD) or Coefficient of Variation (CV) of the distance matrix is one of the most significant features~\cite{smith2010understanding,mersmann2012local,cricsan2021randomness} in algorithm selection or hardness prediction for TSP. Intuitively, when the SD of the TSP distance matrix is very high, it is easy to tell the difference between candidate solutions, and the TSP is easy to solve. At the opposite end of the spectrum, when the SD of the TSP distance matrix is very small, there are many routes with the same minimum cost,  and finding one of them is not difficult. So as the SD increases, an easy-hard-easy transition can be observed~\cite{cheeseman1991really}. Based on the above analysis, 
we add the SD aggregator, along with the MAX aggregator and SUM aggregator, as the three aggregators in our GINES to aggregate useful information for TSP algorithm selection.
\paragraph{\textbf{Extracting local spatial information}} In a TSP instance, cities are distributed in a 2D Euclidean Space. The main characteristic to distinguish TSP instances is the spatial distributions of cities. There exists a research topic that also focuses on learning the spatial distribution of points, namely, point cloud classification. The point cloud is a type of practical 3D geometric data. Identifying point clouds is an object recognition task with many real-world applications, such as remote sensing, autonomous driving, and robotics~\cite{qi2017pointnet++}. Unlike image data made up of regular grids, the point cloud is unstructured data as the distance between neighboring points is not fixed.  As a result, applying the classic convolutional operations on point clouds is difficult. To tackle this, researchers have designed several GNN architectures, such as PointNet\texttt{++}~\cite{qi2017pointnet++}, DGCNN~\cite{wang2019dynamic}, and Point Transformer~\cite{zhao2021point}. 
In the message-passing formulation of these GNNs for point clouds, a common component is $\left(\mathbf{p}_j - \mathbf{p}_i\right) $, here $\mathbf{p}_i$ and $\mathbf{p}_j$ indicate the positions of the current point and neighborhood points, respectively. Through this calculation, local neighborhood information, such as distance and angles between points, can be extracted~\cite{wang2019dynamic}. As the TSP instances can be viewed as 2D point clouds, extracting more local spatial information may help identify the TSP instances' class. We add this component to the message-passing formulation of GINES, as shown follows:
\begin{align}
\mathbf{x}^{\prime}_i = &  \textrm{MLP} \left( (1 + \epsilon) \cdot\mathbf{x}_i + \right. \nonumber\\
& \left. \Box_{j \in \mathcal{N}(i)} \textrm{ReLU}\left(h_{\mathbf{\Theta}}\left(\mathbf{x}_j-\mathbf{x}_i\right) 
+  \mathbf{e}_{j,i}\right)
\right)
\label{XX}
\end{align}
where $\Box$ indicates the selected aggregator, it can be either SD aggregator, MAX aggregator, or SUM aggregator.
$h_{\mathbf{\Theta}}$ is a neural network and defaults to be one linear layer to transform the local spatial information. The whole neural network architecture of our GINES is shown in Figure \ref{fig1}. We adopt three GINES layers to extract the salient spatial information from TSP graphs and apply graph-level Sum pooling for each GINES layer to obtain the entire graph's representation in all depths of the model. Then we concatenate these representations together and feed them into the following two linear layers. We make full use of the learned representation in the first two GINES layers as they may have better feature generalization ability~\cite{xu2018powerful}. 

\begin{figure}[ht]
\centering
\includegraphics[width=0.35\textwidth]{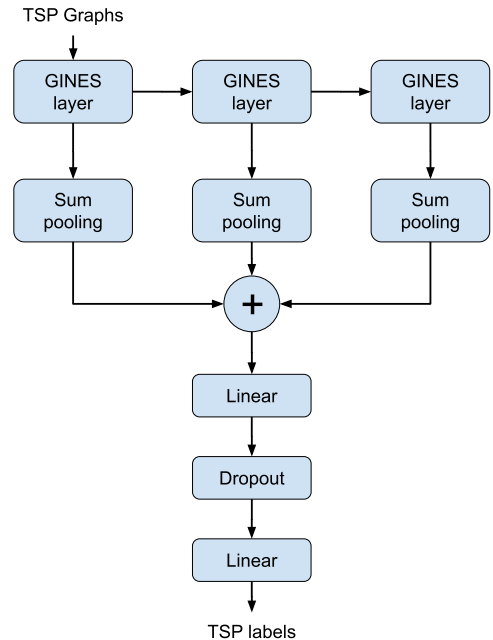}
\caption{The GINES neural network architecture for TSP algorithm selection} \label{fig1}
\end{figure}

\section{Experiments}
\subsection{Dataset}
We evaluate the proposed GINES on two public TSP algorithm selection datasets. The first dataset is generated to assess the Instance Space Analysis (ISA) framework~\cite{smith2011discovering}, and the second is for evaluating the proposed CNN-based selector~\cite{seiler2020deep}. The main difference between the two datasets is the size of the instances. The TSP instances in the first dataset all contain 100 cities, while instances in the second dataset are relatively larger and contain 1000 cities. Applying the proposed model to two different datasets helps us examine its adaptability and compare it with other models. The following part is a detailed description of the two datasets. 
\paragraph{\textbf{TSP-ISA dataset}} includes 1330 TSP instances with 100 cities, and it is divided equally into seven groups based on instance characteristics: RANDOM, CLKeasy, CLKhard, LKCCeasy, LKCChard, easyCLK-hardLKCC and hardCLK-easyLKCC. Here Chained Lin-Kernighan (CLK) and Lin-Kernighan with Cluster Compensation (LKCC) are two well-known local search algorithms for solving TSP. The aim is to predict whether CLK or LKCC is better for each instance, and we can view it as a binary classification task. 
Here LKCC is the Single-Best-Solver, which denotes LKCC can achieve the best average performance across the entire set of problem instances. As the dataset is not balanced, selecting LKCC for all instances can achieve $71.43\%$ accuracy. Here we apply a random oversampling to obtain a balanced training set.
\paragraph{\textbf{TSP-CNN dataset}} includes 1000 TSP instances with 1000 cities. There are two algorithms, one genetic algorithm, Edge-Assembly-Crossover (EAX), and one local search algorithm Lin-Kernighan Heuristic (LKH), to be selected. Here the TSP instances are well-designed to be easy for one algorithm and hard for another.  The TSP-CNN dataset is well-balanced, and choosing EAX, the Single-Best-Solver for all instances, can only achieve $49\%$ accuracy. In addition, the entire dataset is randomly divided into ten folds, allowing us to conduct the same 10-fold cross-validation and make a fair comparison. Researchers generate multiple images to represent TSP instances and apply CNN models to select algorithms~\cite{seiler2020deep}. They have shared the trained CNN model files, and we can load the trained model to obtain the test accuracy of CNNs.

\subsection{Baseline model}
In addition to comparing with the model proposed in other articles, we create several baseline models to be TSP selectors. The baseline models can be divided into two groups: traditional feature-based models and GNNs. We can recognize which representation form and corresponding learning model performs better by comparing baseline models with the proposed GINES.

For traditional feature-based models, we use Random Forest (RF) as the classifier as it performs well in the TSP algorithm selection task \cite{seiler2020deep}. The only difference is the TSP features we choose as the input data. In this work, we evaluate four groups of handcrafted TSP features:

\begin{itemize}
\item All140: All 140 TSP features defined by R package named $salesperson$~\cite{bosseksalesperson}. These features can be divided into 10 groups, including Minimum Spanning Tree (MST) features, kNNG features, Angle features, etc. We also use this package to calculate the following groups of features.
\item Top15: after the feature selection procedure, ~\cite{seiler2020deep} propose the best 15 TSP features for the TSP-CNN dataset. Most of those features are statistical values of strong connected components of kNNG, and others are MST features and Angle features.
\item MST19: all the 19 MST features defined by $salesperson$, are multiple statistical values of MST distance and depth. Here we study the MST features as MST is strongly related to TSP and can be used to solve TSP approximately. Besides, MST features are essential features for algorithm selection according to the previous studies~\cite{seiler2020deep}.
\item kNNG51: all the 51 kNNG features defined by $salesperson$, including statistical values of kNNG distances, as well as the weak/strong connected components of the kNNG.
\end{itemize}

For GNN baseline models, we apply two popular GNNs: GCN and GINE. 
Previous research has tried to apply GCN to TSP algorithm selection but discovered that it performed worse than CNN~\cite{zhao2021towards}. GINE is well-known for its powerful representation learning capabilities, outperforming GCN on a variety of graph-level classification tasks.
Though GINE is not new, to the best of our knowledge, this is the first time that it has been applied to the algorithm selection task.
The architecture and parameter settings of baseline GNN models are the same as those of our GINES. The main modification is replacing the corresponding three GINES layers with GCN and GINE. To analyze the role and performance of aggregators, we test GINES with three different aggregators: MAX (GINES-MAX), SUM (GINES-SUM) and SD (GINES-SD). For a fair comparison with other works, we process the datasets in the same way as
~\cite{seiler2020deep}. On the TSP-ISA dataset, we randomly split the entire dataset into training and test datasets and use 10-fold cross-validation to get the average performance. On the TSP-CNN dataset, we apply exactly the same 10-fold cross-validation in~\cite{seiler2020deep} as the data grouping information was released. We set the hidden channel dimension for the GNN layers to be $32$ and apply PairNorm~\cite{zhao2019pairnorm} after each GNN layer to address the over-smoothing problem. We use an Adam optimizer with a $0.01$ learning rate to reduce Cross Entropy loss and train 100 epochs for each model. We apply the Early-Stopping method with 20 patience in GNN models when dealing with the TSP-CNN dataset and fix all the random seeds to be $41$~\cite{dwivedi2020benchmarking} to ensure the results are reproducible. All experiments were performed on a laptop with Intel Core i7-9750H, and the code is built on Pytorch-geometric~\cite{fey2019fast}. 

\subsection{Result and analysis}

\begin{figure*}[ht]
\centering
\includegraphics[width=0.9\textwidth]{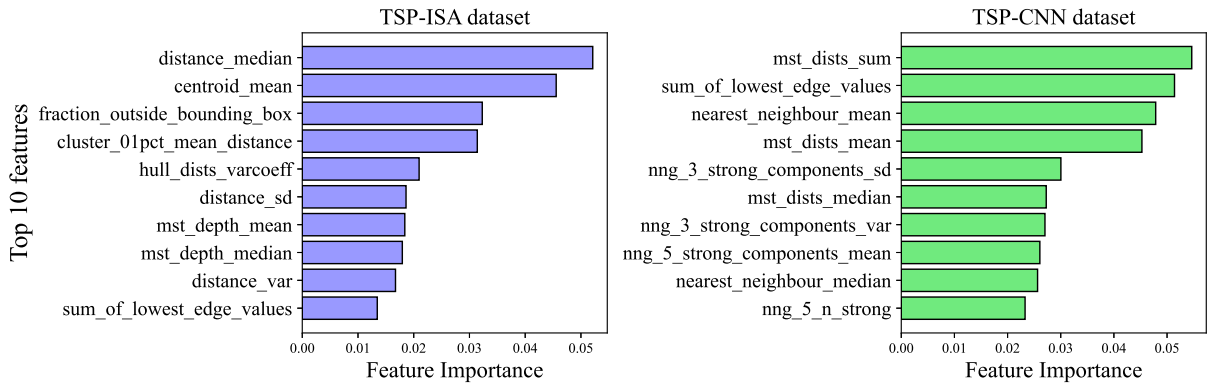}
\caption{The Top 10 importance features for TSP-ISA dataset and TSP-CNN dataset} \label{fig1}
\end{figure*}

The average classification accuracy of each model on the TSP-ISA dataset is listed in Table~\ref{tab1}. The best model's performance is bolded, while the second-best model's performance is underlined. 
We can observe that out of all feature-based approaches, RF with all 140 features performs the best. Employing  fewer features for classification results in substantially lower accuracy. Well-selected features in Top15 can help to keep RF's performance, and MST features are more important than kNNG features in this task. 
GNNs outperform all the traditional feature-based models. GNNs can automatically extract valuable features from kNNG, and they perform much better than  RF with handcrafted kNNG features. Among all GNN architectures, GCN performs relatively poorly compared to GINE and GINES with different aggregators. This result suggests that an elaborate GNN design for this specific application is necessary. By adding a spatial information extractor, our GINES can reach a higher accuracy than the original GINE. We test all three aggregators and the results show that they have comparable performance. As the proposed method does not require any domain knowledge of TSP and has high prediction accuracy, it can be a promising approach in this field. 

\begin{table}[ht]
\renewcommand\arraystretch{1.2}
\caption{Algorithm selection performance comparison on the TSP-ISA dataset.}\label{tab1}
\centering
\setlength{\tabcolsep}{6mm}{
\begin{tabular}{cccccc} \hline
Models  &  Input data  &  Accuracy \\ \hline
\multirow{4}{*}{RF} & All140 features & $95.79\pm2.26$  \\
& Top15 features & $87.37\pm2.42$   \\
& MST19 features & $87.82\pm2.73$   \\
& kNNG51 features & $74.36\pm3.60$   \\ \hline
GCN  & kNNG & $93.38\pm1.71$   \\
GINE  & kNNG & $97.52\pm1.17$   \\
GINES-MAX  & kNNG & $\underline{98.87\pm0.91}$ \\
GINES-SUM  & kNNG & $\textbf{98.87}\pm\textbf{0.61}$  \\
GINES-SD  & kNNG & $98.42\pm0.98$  \\
\hline
\end{tabular}}
\end{table}

\begin{table}[ht]
\renewcommand\arraystretch{1.2}
\caption{Algorithm selection performance comparison on the TSP-CNN dataset.}\label{tab2}
\centering
\setlength{\tabcolsep}{4mm}{
\begin{tabular}{cccccc} \hline
Models  &  Input data  &  Accuracy \\ \hline
\multirow{4}{*}{RF} & All140 features & $73.30\pm5.10$\\
& Top15 features & \underline{$73.40\pm5.66$}\\
& MST19 features & $\textbf{73.90}\pm\textbf{4.81}$ \\
& kNNG51 features & $72.80\pm5.86$   \\ \hline
\multirow{3}{*}{CNN~\cite{seiler2020deep}} & Points$+$MST$+$kNNG images & $70.50\pm7.55$\\
& Points$+$MST images & $72.00\pm4.96$   \\
& Points images & $71.80\pm6.63$   \\ \hline
GCN  & kNNG &  $62.80\pm5.86$ \\
GINE  & kNNG & $66.30\pm3.93$  \\
GINES-MAX  & kNNG & $70.20\pm5.19$ \\
GINES-SUM  & kNNG & $70.00\pm4.47$  \\
GINES-SD  & kNNG & $72.60\pm4.76$ \\
\hline
\end{tabular}}
\end{table}


\begin{table*}[ht]
\renewcommand\arraystretch{1.2}
\caption{Properties comparison between Feature-based model, CNN, and GINES for algorithm selection.}\label{tab3}
\centering
\setlength{\tabcolsep}{5mm}{
\begin{tabular}{cccccc} \hline
Properties  & Feature-based model &  CNN~\cite{seiler2020deep}  &  GINES \\ \hline
\makecell{Intermediate representations preparation}  & Laborious handcrafted features & \makecell{Points images\\MST images\\kNNG images} & None   \\ 
\hline
Need Feature Engineering  & Yes & No & No   \\ 
\hline
\makecell{Problem-irrelevant parameters} & None & \makecell{Image size \\Dot size\\Line width} & None \\
\hline
\makecell{Data Augmentation} & None & \makecell{Random rotation\\Random flipping} & None \\
\hline
\makecell{Problem-relevant information loss} & Affected by domain knowledge & Affected by resolution & None \\
\hline
\makecell{Generalize to VRP\\(Distinguish different points)} & Hard & Hard & \makecell{Easy to add node features} \\
\hline
\makecell{Generalize to ATSP\\(To Non-Euclidean Metric Space)} & Hard & Hard & \makecell{Easy to add edge features} \\
\hline
\end{tabular}}
\end{table*}


The experiment results on the TSP-CNN dataset are shown in Table~\ref{tab2}. Firstly, We apply the feature-based models and find that RF with 
MST features can achieve the best performance. Again, we can observe that MST features are more valuable than kNNG features in the TSP algorithm selection task. Then we load the trained CNN model files and test them to get CNNs' performance. It shows that CNN with Points+MST images is better than CNN with other image inputs. At last, we test the proposed GINES and baseline GNN models. GINES can outperform CNN models but is still worse than feature-based models. The main reason may be the handcrafted features fed into RF are heavily engineered, while the GNN models fail to extract some crucial features, such as MST and clustering features. Besides, there are much more nodes in this dataset, leading to less salient spatial information that can be learned.
 In GINES, we can pick the SD aggregator in the message-passing procedure, because the SD of the distance matrix of TSP is very related to the problem hardness. We also test the prediction accuracy of GINES-MAX and GINES-SUM. The results show that employing an SD aggregator in GINES is a better choice for the TSP-CNN dataset.

Table~\ref{tab3} summarizes the properties of the feature-based model, CNN, and GINES on the TSP algorithm selection task.
Compared to deep learning models such as CNN and the proposed GINES, the traditional feature-based method suffers from the following \emph{shortcomings}. Firstly, substantial domain expertise is required to design features. Secondly, as shown in Figure 2, the important features of the TSP-ISA dataset and the TSP-CNN dataset are significantly different,  indicating that tedious feature engineering is required to choose valuable features. Finally, these selected features are probably inapplicable to other routing problems. The experiment results in Table 2 show that the proposed GINES is a competitive method, and it can slightly outperform CNN in prediction accuracy. GINES has several other advantages compared to CNN. Firstly, CNN takes multiple images as inputs, i.e., Points image, MST image, and kNNG image. Generating these images might be burdensome work, and it is unclear which image can better represent TSP instances. Contrary to CNN, GINES directly takes cities' coordinate and distance matrices as inputs, and we do not need to prepare intermediate representations like images. Secondly, when generating images for CNN, several problem-irrelevant parameters must be set, such as image size, dot size, and line width in MST and kNNG images. Tuning these parameters can be a heavy workload, although theoretically, these parameters should not affect the learned mapping from instances to algorithms. In  GINES, on the other hand, the TSP instances are treated as graphs, and there are not many instance representation parameters to be designed or adjusted. Besides, when setting the image resolution in the CNN method, we should consider the city number in the TSP instance. Otherwise, the representation ability of the image is inadequate, and problem instance information is lost. At last, generating images for TSP instances and applying CNN to select algorithms is not very difficult because cities in TSP are homogeneous and distributed in 2D Euclidean space. If we look into some complex routing problems, we will find that applying the CNN-based method is challenging. For VRP algorithm selection, it is hard to differentiate the depot and customer with image representations. While in GINES, we can simply add the point features to tell them apart. Considering the routing problem in Non-Euclidean space such as ATSP, drawing the problem instance on a 2D plane is nearly impossible. While GINES can naturally recognize the neighborhood in ATSP, we can also modify the message-passing formulation in GINES to aggregate more valuable edge features.

\section{Conclusion}
In this work, we propose a novel GNN named GINES to select algorithms for TSP. By adopting a suitable aggregator and local neighborhood feature extractor, this model can learn useful spatial information of TSP instances and outperform traditional feature-based models and CNNs on public algorithm selection datasets. GINES handles TSP instances as graphs and only takes cities' coordinates and distances between them as inputs. Thus no intermediate representations for problem instances, such as features or images, need to be designed and generated before model training. In contrast to converting TSP instances to images, the graph representation is more natural and efficient, as it neither introduces problem-irrelevant parameters nor loses problem-relevant information. The proposed GINES is promising as it is easy to generalize to other routing problems. For example, we can distinguish nodes and routes in the problem instances by adding node features and edge features. This work can be a good starting point for selecting algorithms or predicting instance hardness for combinatorial optimization problems defined on graphs. In the future, we will explore GINES architectures for more complex problems like ATSP, VRP, and real-world problems. 

\bibliographystyle{IEEEtran}
\bibliography{IEEEabrv,mybibfile}

\end{document}